\documentclass[final,5p,times,twocolumn,authoryear]{elsarticle}

\usepackage{amssymb}
\usepackage{lipsum}
\usepackage{amsmath}
\usepackage{booktabs}

\usepackage{graphicx}

\begin{document}

\begin{frontmatter}

\title{Real-Time Musculoskeletal Surrogates for Pediatric Cerebral Palsy: a Credibility Pilot}

\author[first]{Mohammad Arif Ul Alam}
\affiliation[first]{organization={College of Science and Technology, North Carolina A \& T State University},
            city={Greensboro},
            state={NC},
            country={USA}}

\begin{abstract}
Real-time musculoskeletal (MSK) surrogates could support personalized rehabilitation for children with cerebral palsy (CP), but their credibility depends on subject-wise evaluation, low inference latency, and calibrated uncertainty. We develop a subject-conditioned causal neural surrogate using OpenSim-derived static parameters, temporal joint kinematics, true muscle capacities, and training-only perturbations. On a real pediatric CP gait dataset comprising nine children, we use leave-one-subject-out validation on six development subjects and evaluate a frozen configuration once on three locked test subjects. The surrogate accurately reproduces musculotendon lengths ($R^2\approx0.92$ in development validation and $\approx0.95$ on locked subjects; nRMSE $<8\%$) while requiring only sub-millisecond to few-millisecond neural inference, well below a 100\,ms interactive-rehabilitation target. In contrast, direct muscle-force estimation remains unstable at this small, heterogeneous scale: pooled metrics can overstate within-subject, per-muscle accuracy. A Monte Carlo credibility pilot further shows that propagating only $\pm5\%$ anthropometry and muscle-capacity variation produces severely overconfident nominal 90\% intervals (approximately 4\% force coverage and below 1\% MT-length coverage). These results establish a leakage-free evaluation and credibility framework for pediatric MSK surrogates, while identifying force modeling and epistemic uncertainty as the central next challenges for clinically credible digital twins.
\end{abstract}

\begin{keyword}
Pediatric cerebral palsy, Musculoskeletal modeling, Neural surrogates, Real-time rehabilitation, Credibility assessment
\end{keyword}

\end{frontmatter}

\section{Introduction}

Cerebral palsy (CP) is the leading cause of childhood physical disability and is associated with highly heterogeneous movement impairments \cite{Rosenbaum2007CP}. In particular, crouch gait changes joint loading, muscle recruitment, and the mechanical demands of walking, making patient-specific analysis important for clinical decision-making \cite{Steele2010Crouch}. Musculoskeletal (MSK) simulation provides a noninvasive way to estimate quantities that cannot be measured directly during gait, including muscle forces, activations, and joint moments. OpenSim and subject-scaled full-body models have made these analyses broadly reproducible and clinically relevant \cite{Seth2018OpenSim,Rajagopal2016FullBody}. However, conventional muscle-driven simulations remain too computationally expensive for interactive rehabilitation or continuously updated digital twins.

Neural surrogates offer a practical route to real-time MSK analysis. Recent studies have used temporal neural networks and physics-informed objectives to estimate muscle forces or joint kinematics from motion and electromyography signals \cite{Zhang2023PhysicsInformed,Liu2024CNNLSTM,Zhao2023KnowledgeTransfer}. These results are promising, but most evidence is based on healthy adults, a small number of muscles, or task-specific datasets. Pediatric CP creates a more difficult setting because gait patterns, body dimensions, strength, and neuromuscular control vary substantially between children. Further, muscle force is not directly observed: it is an underdetermined model estimate whose value depends on the musculoskeletal model, optimization method, and physiological assumptions \cite{Roelker2020Optimization,Falisse2018OpenSimHumanBody}. Consequently, a surrogate should be evaluated not only by a favorable frame-pooled score, but also by its ability to generalize across held-out children.

Credibility is equally important for clinical digital twins. ASME V\&V~40 and recent FDA guidance emphasize that the evidence supporting a computational model should be appropriate for its intended context of use \cite{FDA2023Credibility}. This requires transparent data partitioning, validation on independent subjects, and explicit assessment of uncertainty rather than relying on point predictions alone. In this preliminary study, we develop a real-time, subject-conditioned MSK surrogate for pediatric CP gait and contribute: (i) a leakage-free evaluation protocol with leave-one-subject-out development validation and a locked held-out test set; (ii) a causal surrogate conditioned on OpenSim-derived static parameters and true muscle capacities; (iii) an honest characterization of the limitations of small-cohort muscle-force estimation; and (iv) a Monte Carlo credibility pilot showing that input-parameter propagation alone does not yield calibrated uncertainty. The subject-level cohort and evaluation partition are summarized in Table~\ref{tab:cohort}. Together, these results establish a rigorous experimental foundation for future credible pediatric MSK digital twins.

\begin{table}[t]
\centering
\caption{Cohort and subject-level evaluation partition. Frame counts denote
aligned kinematic/CMC ground-truth frames. Dev: development LOSO; Test: locked
held-out evaluation.}
\label{tab:cohort}
\scriptsize
\setlength{\tabcolsep}{3.5pt}
\begin{tabular}{lccc}
\toprule
Subject & Severity & Frames & Split \\
\midrule
MI01 & Mild     & 674   & Dev \\
MI02 & Mild     & 893   & Dev \\
MO02 & Moderate & 799   & Dev \\
MO03 & Moderate & 749   & Dev \\
SE01 & Severe   & 838   & Dev \\
SE02 & Severe   & 889   & Dev \\
\midrule
MI03 & Mild     & 736   & Test \\
MO04 & Moderate & 1,166 & Test \\
SE05 & Severe   & 804   & Test \\
\midrule
Total & --- & 7,548 & 6 Dev / 3 Test \\
\bottomrule
\end{tabular}
\end{table}

\section{Method}
\label{sec:method}

\subsection{Subject-Conditioned Causal Surrogate}

We developed a compact causal temporal convolutional network (TCN) with approximately
0.9 million trainable parameters. TCNs provide a computationally efficient mechanism
for using a finite history of motion while preserving causality, making them appropriate
for real-time sequence inference \cite{Bai2018TCN}. At time $t$, the model receives a
251-frame ($\approx$250\,ms) history of 23 joint coordinates and their speeds
(46 dynamic features), together with subject-specific static parameters extracted from
the scaled OpenSim model. These parameters include muscle-specific maximum isometric
forces ($F^{\max}$), body-scale factors, model mass, and muscle geometric and
contractile properties. OpenSim provides a transparent basis for these subject-scaled
mechanistic quantities \cite{Seth2018OpenSim,Rajagopal2016FullBody}.

The dynamic sequence is encoded by the causal TCN, while a small multilayer perceptron
encodes static parameters. Their fused representation produces muscle force,
activation, musculotendon (MT) length, and moment-arm outputs:
\begin{equation}
\begin{aligned}
(\hat{\mathbf f}_t,\hat{\mathbf a}_t,\hat{\boldsymbol{\ell}}_t,
\hat{\mathbf R}_t)
&= g_{\theta}(\mathbf X_{t-W+1:t},\mathbf s), \\
\hat{\boldsymbol{\tau}}_t
&= \hat{\mathbf R}_t^{\top}\hat{\mathbf f}_t, \qquad
\hat{\mathbf f}_t =
\mathbf F^{\max}\odot \hat{\mathbf a}_t\odot
\sigma(\mathbf z_t),
\end{aligned}
\label{eq:surrogate}
\end{equation}
where $W=251$, $\mathbf X$ denotes kinematics, $\mathbf s$ denotes static
subject parameters, $\mathbf R$ contains muscle moment arms, and
$\boldsymbol{\tau}$ is the resulting net muscle joint moment. The final expression
enforces a muscle-specific force-capacity bound using true OpenSim
$F^{\max}$ values. The model predicts 92 muscle forces, activations, and MT lengths,
as well as moment arms for 23 independent coordinates.

\subsection{Supervision and Physics-Informed Objective}

Ground-truth forces and activations were obtained from the existing OpenSim computed
muscle control (CMC) workflow, and MT lengths, moment arms, and inverse-dynamics joint
moments were generated from each subject-scaled model. This use of OpenSim-derived
labels follows established musculoskeletal simulation practice
\cite{Thelen2003CMC,Seth2018OpenSim}. We optimized a multi-task objective:
\begin{equation}
\mathcal{L} =
\lambda_f\mathcal{L}_{f}+
\lambda_a\mathcal{L}_{a}+
\lambda_{\ell}\mathcal{L}_{\ell}+
\lambda_R\mathcal{L}_{R}+
\lambda_{\tau}\mathcal{L}_{\tau},
\end{equation}
where each term is a normalized mean-squared error for force, activation, MT length,
moment arm, or joint moment, respectively. The torque term is evaluated only on
available lower-limb coordinates:
\begin{equation}
\mathcal{L}_{\tau} =
\frac{1}{|\mathcal J|}
\sum_{j\in\mathcal J}
\left(
\hat{\tau}_{t,j}-\tau^{ID}_{t,j}
\right)^2 .
\end{equation}
Thus, predicted muscle forces are coupled to joint-level mechanics through the
moment-arm relation in Eq.~\ref{eq:surrogate}. Physics-informed learning has shown
promise for musculoskeletal inference, but its benefit depends on label quality,
modeling assumptions, and the available data regime \cite{Zhang2023PhysicsInformed}.
Accordingly, we report a direct ablation in Section~\ref{sec:results}, rather than
presuming that this term improves force estimation.

\subsection{Training-Only Perturbations}

To improve robustness without contaminating evaluation, we applied synthetic
perturbations exclusively to training windows. These included independent
$\pm5\%$ perturbations of $F^{\max}$ and anthropometry-related static features,
small additive kinematic sensor noise, and temporary occlusion of up to two input
channels. Validation and locked-test windows remained unperturbed. This separation is
important because augmentation must not create an implicit pathway from evaluation
subjects into model selection.

\subsection{Leakage-Free Evaluation Protocol}

Our primary methodological contribution is a subject-level evaluation protocol. The
six development subjects were evaluated with leave-one-subject-out (LOSO) validation:
five subjects were used for training and the remaining subject for validation in each
fold. Dynamic and static normalizers, missing-value imputations, augmentation, and
early-stopping decisions were fitted using only the corresponding training subjects.
Three additional children, one per crouch-severity group, were retained as a locked
held-out test set. The final architecture and training duration were frozen from the
development folds before this test set was evaluated once.

\begin{figure*}[t]
    \centering
    \includegraphics[width=\textwidth]{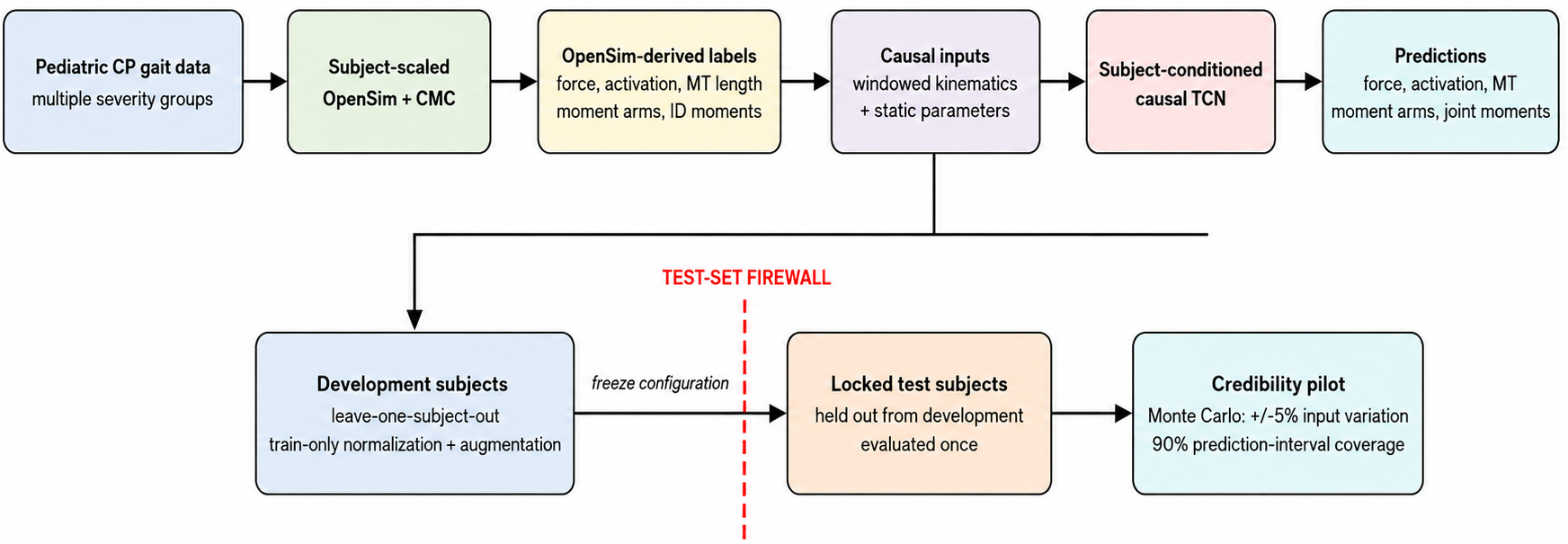}
    \caption{Overview of the subject-conditioned musculoskeletal surrogate and
    leakage-free evaluation protocol. Pediatric CP gait data are processed with
    subject-scaled OpenSim models and computed muscle control to generate force,
    activation, musculotendon-length, moment-arm, and inverse-dynamics labels.
    Windowed kinematics and static model parameters are provided to a causal TCN.
    Development subjects are used exclusively for leave-one-subject-out model
    selection and training-only augmentation. A firewall separates the locked
    held-out evaluation from development. The frozen model is subsequently used
    for a Monte Carlo credibility pilot with $\pm5\%$ input variation and nominal
    90\% prediction-interval coverage assessment.}
    \label{fig:method}
\end{figure*}

We report within-subject performance as the primary measure, including subject-level
and mean-per-muscle $R^2$ values. Frame-pooled/global $R^2$ is reported only alongside
these measures because it can be inflated by between-subject force-scale differences.
The complete surrogate, data-labeling, test-firewall, and credibility-pilot workflow is
shown in Figure~\ref{fig:method}. The protocol and uncertainty screening pilot are intended
to support a risk-informed credibility workflow consistent with the principles of ASME
V\&V~40 and current FDA guidance for computational modeling and simulation
\cite{FDA2023Credibility}.

\section{Results}
\label{sec:results}

\subsection{Musculotendon Geometry Surrogate}

The causal subject-conditioned surrogate reproduced musculotendon (MT) lengths
accurately. Development leave-one-subject-out (LOSO) validation achieved pooled
MT-length $R^2=0.9245$, while the frozen model achieved $R^2=0.9477$ on the
locked held-out subjects (Table~\ref{tab:results}; Figure~\ref{fig:subject-results}).
Locked-test MT-length nRMSE was 7.69\% overall, with values of 6.51\%, 6.38\%,
and 9.51\% for the mild, moderate, and severe held-out subjects, respectively.

This result establishes that the pipeline can replace the OpenSim geometry mapping
with high accuracy in a causal setting. However, MT length is a smooth,
largely deterministic function of joint configuration in the scaled musculoskeletal
model \cite{Seth2018OpenSim,Thelen2003CMC}. Thus, this result should be interpreted
as validation of the data alignment, subject conditioning, and real-time geometry
surrogate, rather than as evidence that MT-length prediction is the most difficult
scientific component of the problem.

\subsection{Real-Time Inference}

For one causal input window, mean neural inference time was 7.41\,ms on a
single-thread CPU and 3.16\,ms on an NVIDIA A100 GPU. Both measurements are
substantially below the 100\,ms interactive-rehabilitation criterion. The result
therefore establishes real-time feasibility for the neural surrogate component.

We do not report a numerical speedup relative to an OpenSim CMC or static-optimization
solve because an equivalent end-to-end OpenSim latency measurement was not collected
in the current notebook. Such a comparison should be added in a follow-up experiment;
otherwise, an explicit orders-of-magnitude speedup claim would not be supported by the
present evidence.

\subsection{Direct Muscle-Force Estimation Remains Difficult}

\begin{figure*}[htbp]
    \centering
    \includegraphics[width=\textwidth]{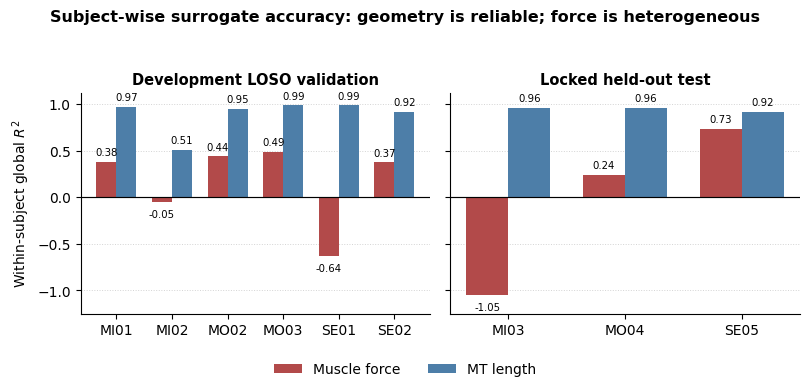}
    \caption{Subject-wise global $R^2$ for muscle force and MT length.
    Development subjects are evaluated by leave-one-subject-out validation;
    locked subjects are evaluated once after model configuration freezing.
    MT-length prediction is consistently strong except for one development
    subject, whereas force prediction varies sharply across children. This
    variation demonstrates why pooled force $R^2$ must be interpreted alongside
    subject-wise metrics.}
    \label{fig:subject-results}
\end{figure*}

Force prediction was substantially less stable than MT-length prediction. Across
development LOSO folds, pooled force $R^2$ was 0.2165 with nRMSE of 34.55\%.
Subject-level force $R^2$ ranged from $-0.635$ to 0.486. On the locked test set,
the pooled force $R^2$ increased to 0.6056, but this pooled value masked pronounced
heterogeneity: force $R^2$ was $-1.052$ for MI03, 0.239 for MO04, and 0.733 for
SE05. A negative value means that the model performed worse than a subject-specific
mean-force predictor. The complete subject-wise force and MT-length results are shown
in Table~\ref{tab:results} and Figure~\ref{fig:subject-results}.

This discrepancy illustrates why pooled force metrics must not be used as the primary
claim in a small, heterogeneous pediatric cohort. Frame-pooled/global $R^2$ can partly
reflect between-subject differences in force scale rather than within-subject temporal
tracking. We therefore emphasize subject-wise results in Figure~\ref{fig:subject-results}
and recommend reporting normalized force, $f/F^{\max}$, in future experiments as a
scale-robust secondary outcome. This normalized metric was not computed in the current
notebook and is not retrospectively claimed here. Overall, the kinematic and latency
components of the feasibility hypothesis are supported, whereas the direct muscle-force
accuracy target is not yet met.

\subsection{Torque-Consistency Diagnostic}

The physics-informed torque pathway did not yield meaningful joint-moment accuracy in
the present configuration. Development LOSO pooled joint-moment $R^2$ was $-0.0055$,
and subject-wise values ranged from $-0.024$ to 0.000
(Table~\ref{tab:results}). This indicates that the masked torque-consistency term was
empirically inert at the available sample size and label quality.

This finding does not invalidate physics-informed musculoskeletal learning; rather, it
shows that including a mechanically motivated loss does not guarantee improved
generalization \cite{Zhang2023PhysicsInformed}. A future formal ablation should compare
otherwise identical models trained with and without the torque term, and should evaluate
normalized force targets, Hill-type force generation, and improved joint-moment labels.

\begin{table}[t]
\centering
\caption{Subject-wise global $R^2$ results. Each subject-level value is computed
over that subject's evaluation windows and output dimensions. Joint-moment labels
were generated only for development LOSO evaluation.}
\label{tab:results}
\scriptsize
\setlength{\tabcolsep}{3pt}
\begin{tabular}{llrrr}
\toprule
Set & Subject & Force & MT length & Joint moment \\
\midrule
Dev. & MI01 & 0.380 & 0.974 & -0.019 \\
Dev. & MI02 & -0.050 & 0.512 & -0.024 \\
Dev. & MO02 & 0.441 & 0.951 & 0.000 \\
Dev. & MO03 & 0.486 & 0.987 & -0.014 \\
Dev. & SE01 & -0.635 & 0.986 & -0.013 \\
Dev. & SE02 & 0.374 & 0.919 & -0.007 \\
\midrule
Dev. & Pooled & 0.217 & 0.925 & -0.006 \\
\midrule
Locked & MI03 & -1.052 & 0.962 & --- \\
Locked & MO04 & 0.239 & 0.961 & --- \\
Locked & SE05 & 0.733 & 0.919 & --- \\
\midrule
Locked & Pooled & 0.606 & 0.948 & --- \\
\bottomrule
\end{tabular}
\end{table}

\section{A Credibility (VVUQ) Pilot}
\label{sec:vvuq}

We conducted a screening-level uncertainty-propagation experiment using the frozen
locked-test model. For each held-out subject, we drew 100 Monte Carlo samples with
independent $\pm5\%$ perturbations of muscle-specific $F^{\max}$ values and
anthropometry-related static inputs. We evaluated nominal 90\% prediction intervals
over randomly sampled causal windows and measured empirical coverage and mean interval
width. This experiment is a sensitivity and calibration diagnostic, not a complete
credibility assessment.

\begin{figure*}[htbp]
    \centering
    \includegraphics[width=\textwidth]{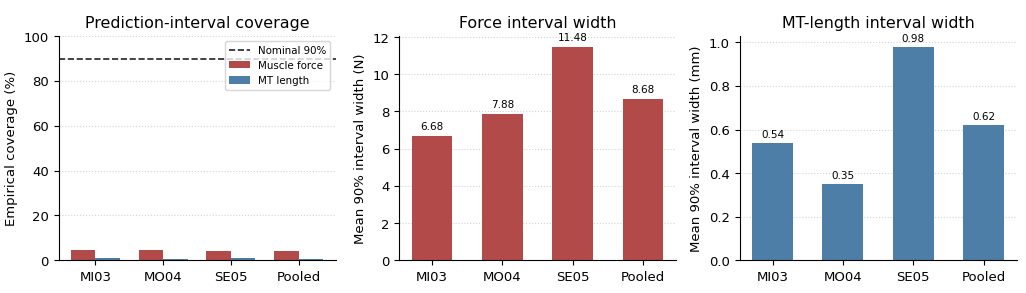}
    \caption{Monte Carlo credibility pilot on locked held-out subjects.
    The dashed line denotes nominal 90\% coverage. Propagating only $\pm5\%$
    variation in muscle capacity and anthropometry-related static inputs yields
    severely overconfident intervals: pooled empirical coverage is 4.35\% for
    force and 0.70\% for MT length. Narrow interval widths show that
    input-parameter variation captures only a small component of total
    predictive uncertainty.}
    \label{fig:vvuq}
\end{figure*}

The resulting intervals were severely overconfident (Figure~\ref{fig:vvuq}). Across
the locked cohort, nominal 90\% intervals achieved only 4.35\% empirical coverage for
muscle force and 0.70\% coverage for MT length. Mean interval widths were 8.68\,N for
force and 0.00062\,m for MT length. The same pattern appeared for every held-out
subject: force coverage ranged from 3.94\% to 4.72\%, and MT-length coverage ranged
from 0.37\% to 0.88\%. Thus, the propagated input variation represents only a small
fraction of the observed prediction error.

This failure is itself an informative result. The dominant uncertainty source is likely
model-form and epistemic uncertainty associated with limited, heterogeneous pediatric
training data, rather than the narrow input-parameter variation considered here.
Consequently, input-only propagation is insufficient for clinically consequential or
regulatory-grade credibility claims. Future work will add epistemic uncertainty through
deep ensembles \cite{Lakshminarayanan2017DeepEnsembles}, heteroscedastic prediction
heads, and calibration methods such as conformal prediction
\cite{Angelopoulos2021Conformal}. In the spirit of ASME V\&V~40, this pilot establishes
a measurable credibility gap and identifies the additional validation evidence needed
before the surrogate could support a higher-consequence decision
\cite{FDA2023Credibility}.

\section{Discussion and Limitations}
\label{sec:discussion}

This preliminary study establishes a rigorous evaluation substrate for pediatric MSK surrogates rather than a claim that all components of a clinically deployable digital twin have been solved. The leakage-free subject protocol, locked-test firewall, train-only preprocessing, and fast MT-length predictions demonstrate that real pediatric CP data can support reproducible, real-time surrogate evaluation. As summarized in Table~\ref{tab:results} and Figure~\ref{fig:subject-results}, however, the heterogeneous force results show that direct regression of raw muscle force from kinematics remains insufficient at this exploratory scale. In particular, pooled force metrics should not substitute for within-subject, per-muscle performance because body size and force scale can inflate frame-pooled agreement.

Several limitations define the next research steps. First, the cohort contains only nine children and one lower-limb gait trial per subject, as summarized in Table~\ref{tab:cohort}; larger and more diverse pediatric datasets are required to model inter-subject heterogeneity and to support reliable subgroup analyses. Second, CMC-derived force and activation labels are physics-based model estimates rather than in-vivo measurements. Future work should therefore predict normalized force ($f/F_{\max}$), incorporate Hill-type muscle dynamics, and evaluate sensitivity to alternative optimization and muscle-model assumptions \cite{Thelen2003CMC,Seth2018OpenSim}. Third, the present torque-consistency pathway did not provide measurable improvement, indicating that moment supervision, moment-arm estimation, loss weighting, or the force parameterization requires redesign rather than merely stronger regularization.

Finally, the credibility pilot in Figure~\ref{fig:vvuq} demonstrates that propagating plausible input uncertainty is not equivalent to calibrated predictive uncertainty. Regulatory-grade credibility requires explicit treatment of model-form and epistemic uncertainty, for example through ensembles, heteroscedastic prediction, and conformal calibration \cite{Lakshminarayanan2017DeepEnsembles,Angelopoulos2021Conformal,FDA2023Credibility}. These findings directly motivate the next phase of the program: expansion to additional pediatric subjects, improved physics-consistent force modeling, calibrated uncertainty quantification, and extension from lower-limb gait to upper-limb rehabilitation tasks.

\section{Conclusion}
\label{sec:conclusion}

We developed a leakage-free, subject-conditioned MSK surrogate pipeline for real pediatric CP gait data. The approach meets real-time latency and MT-length accuracy targets, while quantifying the unresolved difficulty of small-cohort muscle-force estimation and showing that input-only uncertainty propagation is severely overconfident. Together, these results, summarized in Table~\ref{tab:results} and Figures~\ref{fig:subject-results}--\ref{fig:vvuq}, provide a credible preliminary foundation and a focused technical agenda for in-silico pediatric neurorehabilitation.

\bibliographystyle{elsarticle-harv} 
\bibliography{example}

\end{document}